# A Study on the Implementation of Generative AI Services Using an Enterprise Data-Based LLM Application Architecture

Cheonsu Jeong[1*]

[1] Principal Consultant & the Technical Leader for AI Automation Platform at SAMSUNG SDS
125, Olympic-ro 35-gil, Songpa-gu, Seoul 05510, Korea.

**Abstract**

This study presents a method for implementing generative AI services by utilizing the Large Language Models (LLM) application architecture. With recent advancements in generative AI technology, LLMs have gained prominence across various domains. In this context, the research addresses the challenge of information scarcity and proposes specific remedies by harnessing LLM capabilities. The investigation delves into strategies for mitigating the issue of inadequate data, offering tailored solutions. The study delves into the efficacy of employing fine-tuning techniques and direct document integration to alleviate data insufficiency. A significant contribution of this work is the development of a Retrieval-Augmented Generation (RAG) model, which tackles the aforementioned challenges. The RAG model is carefully designed to enhance information storage and retrieval processes, ensuring improved content generation.

The research elucidates the key phases of the information storage and retrieval methodology underpinned by the RAG model. A comprehensive analysis of these steps is undertaken, emphasizing their significance in addressing the scarcity of data. The study highlights the efficacy of the proposed method, showcasing its applicability through illustrative instances. By implementing the RAG model for information storage and retrieval, the research not only contributes to a deeper comprehension of generative AI technology but also facilitates its practical usability within enterprises utilizing LLMs. This work holds substantial value in advancing the field of generative AI, offering insights into enhancing data-driven content generation and fostering active utilization of LLM-based services within corporate settings.

**Keywords:** Generative AI, LLM, RAG, LangChain, LLM framework, Embedding, Vector Store

---

[*] Corresponding author and first author: csu.jeong@samsung.com

# I. Introduction

Recent developments in generative AI, catalyzed by ChatGPT, have become a focal point of discussion. Generative AI possesses the potential to contribute across a myriad of domains, encompassing natural language generation, translation, and the generation of diverse and imaginative content. A notable indication of global anticipation for generative AI is underscored by the '2023 Emerging Technologies Hype Cycle' report by Gartner, released on August 17, 2023. Within the realm of 'Emergent AI,' which denotes burgeoning artificial intelligence, the category of 'generative AI' is marked as being at the 'Peak of Inflated Expectations.' The report forecasts that within 2 to 5 years, the field of generative AI will achieve transformative accomplishments, heralding a new era of enhanced human productivity and machine creativity.

Of particular significance, Large Language Models (LLM) such as OpenAI's GPT series have demonstrated groundbreaking outcomes in the realm of natural language comprehension and generation. The wave of generative AI, instigated by ChatGPT, has extended its influence to encompass visual mediums, including Stable Diffusion and Midjourney, capturing the attention of the general populace.

In light of these advancements, LLM are being extensively harnessed across diverse domains. Leveraging extensive training on copious volumes of textual data, LLM exhibit the ability to comprehend and generate natural language. Building upon this prowess, LLMs find application in various domains, including customer interactions, creative content creation, and question-answering.

However, generative AI still faces various limitations. LLM require extensive amounts of data for training, which incurs substantial costs and time investment. Furthermore, LLMs exhibit limited adaptability to new data, making it challenging to provide accurate responses to questions unrelated to the data they were trained on. Notably, ChatGPT, provided by OpenAI, tends to exhibit a phenomenon known as "hallucination." This entails fabricating information when faced with queries about unfamiliar facts. Consequently, while responses may appear plausible on the surface, they often contain incorrect information. In efforts to mitigate this, approaches to minimize hallucination and elicit responses that align with actual data are being pursued. Strategies include appending context to prompts, employing Chain-Of-Thought (CoT) techniques, enhancing self-consistency, and requesting concise answers from the model.

Another constraint arises from the limited answer capacity of LLMs due to information gaps. For instance, the GPT 3.5 model lacks data beyond September 2021, rendering it incapable of furnishing responses pertaining to recent news events. Moreover, since external information is utilized for generating answers, there is a growing demand for methods to address inquiries related to sensitive business insider information.

As a primary solution to address these challenges, fine-tuning of LLM with new data is proposed. This approach involves additional training on specific domains using fresh datasets. For ultra-large AIs, updating all parameters can be arduous. Consequently, strategies such as LoRA and P-tuning are employed, focusing on training only a subset of parameters. In pursuit of overcoming such limitations, OpenAI introduced the capability to fine-tune the GPT-3.5 Turbo model, a significant advancement unveiled in August 2023. Furthermore, plans are in place to offer fine-tuning capabilities for the GPT-4 model by the autumn of 2023. This update empowers developers to customize models according to their use cases, harnessing enhanced performance. These tailored models can subsequently be deployed at scale.

Initial tests have demonstrated that the fine-tuned GPT-3.5 Turbo version exhibits comparable, and in some cases even superior, performance to the base GPT-4 model across specific narrow tasks. Notably, similar to all APIs, data transmitted through the fine-tuning API is owned by the customers, and OpenAI or any other organization is precluded from employing this data for training alternative models, as officially stated.

However, this approach entails significant costs. Currently, the fine-tuning expenses for the GPT-3.5 Turbo model are categorized into two primary buckets: initial training costs and usage costs. For instance, the anticipated cost for a gpt-3.5-turbo fine-tuning task involving a training file containing 100,000 tokens trained over three epochs is approximately $2.40. An alternative strategy involves directly incorporating user-desired information, encapsulated within documents, into the prompt context. Nonetheless, the practicality of manually embedding all information into the context is unfeasible. GPT-3.5 has the capacity to store information equivalent to roughly 8,000 words or around 5 pages of text, while GPT-4's processing capabilities are restricted to inputs of up to approximately 50 pages. In such scenarios, it is more efficient for users or businesses to store their information in databases. Subsequently, when user queries arise—such as inquiries regarding company dress codes through a chatbot—pertinent information can be retrieved and presented to the LLM through prompts, proving to be a more practical and efficient approach. For instance, the approach involves uploading a PDF document and posing a question, whereby the system searches for relevant information within the PDF to provide an answer. This methodology represents the second approach, termed Retrieval Augmented Generation (RAG) service architecture. In this manner, RAG informs the LLM of pertinent queries and associated reference materials in advance, mitigating hallucination tendencies and enabling more accurate response generation. The RAG architecture thus addresses the information scarcity issue within LLM, possessing the potential to furnish high-quality responses without necessitating new data training. The RAG model offers a means to supply users with more precise and fitting answers to their inquiries, making it a valuable technology for real-world business contexts. This versatility renders it applicable across various business domains, enhancing LLM performance. Efforts to enhance language model performance through leveraging search techniques have long been underway. Notably, in 2021, DeepMind introduced RETRO, utilizing its internal database for information retrieval, while OpenAI unveiled WebGPT with Bing-based searching capabilities in the same year. However, the current prominence of the RAG architecture in the industry stems from its remarkable improvement in in-context learning abilities and the convenience of not requiring separate model training efforts.

In this manner, addressing the limitations of ChatGPT and harnessing internal information entails avenues such as fine-tuning through additional training with new data specific to certain domains, or employing the RAG technique. Through RAG, pre-defined queries and associated reference materials are supplied to the LLM, enhancing the accuracy and reliability of ChatGPT. However, contemporary solutions to these challenges lack readily available real-world implementation instances or methodologies.

Against this backdrop, this study seeks to propel the advancement of generative AI while surmounting its limitations, by exploring methods to implement LLM applications using the RAG architecture. It proposes procedures and methodologies to facilitate the facile implementation of such applications and presents illustrative implementation cases. To this end, the study initially examines approaches to overcome the information scarcity of LLM through fine-tuning or direct utilization of document information. It subsequently delves into the operational mechanics and key phases of the RAG model, discussing methods of information storage and retrieval using vectorized databases. Furthermore, it encompasses an explanation of furnishing suitable prompts to LLM and the orchestration framework for the same. Consequently, the study introduces specific implementation methods and available tools, comprehensively detailing the process of realizing the RAG model. Contemplating the potential of contemporary LLM models and vectorized database technologies, it explores the amalgamation and optimization of these technologies. Additionally, it presents implementation codes for the RAG model across diverse business domains, analyzing insights and outcomes derived from these endeavors. In doing so, the study deliberates upon the practical applicability and potential benefits of the RAG model, thereby enhancing generative AI services and discerning avenues for advancement. Furthermore, this research contributes by providing implementation codes leveraging generative AI technology, investigating the feasibility of real-world business application, and fostering the development and industrial utilization of generative AI technology.

In Chapter 2, the theoretical foundation is established by detailing prior research and key related concepts pertinent to this study. The exploration encompasses the trends in generative AI and LLM on both domestic and international fronts. It further elucidates strategies to overcome the information scarcity inherent in LLM, a prerequisite for implementing the RAG model. In Chapter 3, a concrete implementation methodology is delineated. This involves explicating the procedure to implement the RAG model employing a variety of open-source and commercial tools. A comparative analysis between conventional data fine-tuning and direct document information utilization is conducted, evaluating their respective merits and demerits. Subsequently, an optimal approach is proposed, considering cost-effectiveness. The core aspects of the RAG model—information retrieval and utilization—are exhaustively investigated. The process of embedding document information using vectorized databases, its storage, and subsequent retrieval is elaborated. An innovative method is introduced to convert retrieval outcomes into prompts for LLM through an orchestration framework, facilitating the effective synthesis of user queries and search results to generate optimal responses. In Chapter 4, the application of the RAG model in various business domains is substantiated through the presentation of actual implementation codes, utilizing contemporary LLM models and vectorized DB technology. This serves to affirm the practical feasibility of generative AI technology. Finally, Chapter 5, the conclusion, encapsulates the research findings, identifies limitations, and outlines future research directions.

## II. Related Work

For the purpose of this study, an extensive investigation into recent significant research papers, journals, articles, and books related to generative AI and LLM has been conducted. This chapter delves into a comprehensive exploration of both LLM and generative AI as a whole. It commences by delving into the concept and diverse application domains of generative AI. Furthermore, it examines the intricate elements, technological components, and frameworks pertinent to LLM, as well as demarcates the realm of RAG that is the focus of this paper.

In this chapter, an exhaustive review of pertinent literature has been undertaken to contextualize the research within the contemporary landscape of generative AI and LLM. The chapter unfolds by providing a detailed understanding of the concepts and applications of generative AI, followed by a meticulous breakdown of the technical components and frameworks underlying LLM, and concludes by delineating the RAG domain.

Through this literature review, we aim to establish a comprehensive understanding of the theoretical foundations and recent advancements in the field, setting the stage for the subsequent chapters that delve into the practical implementation and evaluation of the proposed RAG model.

### 2.1 Background for Generative AI

Generative AI is a form of artificial intelligence that utilizes extensive trained data models to generate new content such as text, images, audio, and videos. Prominent examples in this domain include ChatGPT, which is a language model service trained on vast amounts of data, and models like DALL-E and Midjourney, which focus on generating images. The categorization of generative AI models varies based on their output generation, distinguishing them as language models, image models, video models, and so on. However, the landscape is rapidly evolving towards multi-modal models that can learn both text and images simultaneously, positioning themselves as foundation models.

In the context of AI, IDC positions generative AI as illustrated in Fig 1., encompassing unsupervised and semi-supervised algorithms that enable computers to respond to short prompts and create new content using previously generated content such as text, audio, video, images, and code [1]. Over the past few years, generative AI technology has undergone rapid advancement, ushering in novel opportunities across various industries.

Unlike traditional methods that merely process or analyze existing data, generative AI employs a novel approach to generate fresh and inventive content. These models learn patterns and are trained on extensive datasets to generate new outputs resembling the training data.

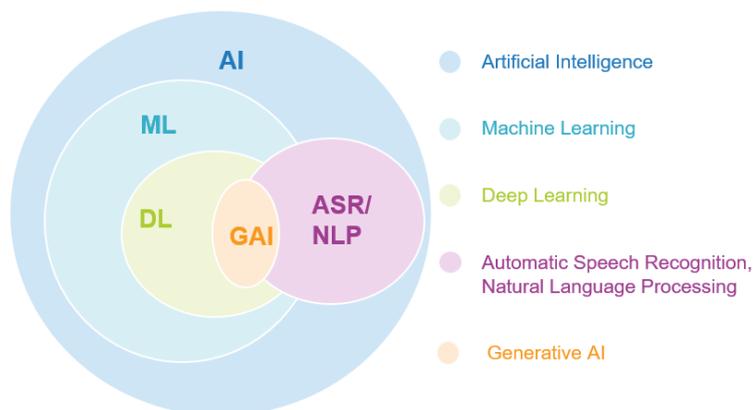

**Fig 1** Generative AI Relation Diagram (IDC, 2023)

2.1.1 Generative AI types and applications

Generative AI technology has the remarkable ability to create a wide range of data forms, including text, code, images, videos, 3D models, and audio. Table 1. showcases representative models within generative AI technology based on different data formats and the application domains they serve [2, 3].

**Table 1** Representative models and applications of generative AI

| Type | Representative Models | Application Field |
| --- | --- | --- |
| Text | OpenAI GPT-4, Google PaLM2, DeepMind Gopher, Meta LLaMA, Hugging Face Bloom | Marketing, Sales, Customer Support, General Writing, Translation, Summarization |
| Code | OpenAI Codex, Google PaLM2, Ghostwriter, Amazon CodeWhisperer, Tabnine, Stenography, AI2sql, Pygma | Code Generation, Document Code, Generate SQL Queries, Generate Web Apps, Testing, Code formatting |
| Image | OpenAI Dall-E2, Stable Diffusion, Midjourney, Google Imagen, Meta Make-A-Scene | Generate Image, SNS, Advertising, Design, Data visualization |
| Video | MS X-CLIP, Meta Make-A-Video, RunwayML, Synthesia, Rephrase AI, Hour One | Video Generation, Video Editing, , Video Summarization |
| 3D | DreamFusion, NVIDIA GET3D, MDM | Generate 3D Models, Generate 3D Scenes |
| Audio | Resemble AI, WellSaid, Play.ht, Coqui, Harmonai, Google MusicLM | Speech Synthesis, Voice Cloning, Generate Music, Sound effect design |

For this study, we investigated materials related to generative AI and LLMs, including recent major research papers, journals, articles, and books. In this chapter, we will explore LLMs and generative AI in general. First, we will learn in detail about the concept and application areas of generative AI. Then, we will explain the key technologies and frameworks that are applied to LLMs in this paper, and the RAG areas.

To conduct this study, we investigated materials related to generative AI and LLMs, including recent major research papers, journals, articles, and books. In this chapter, we will provide an overview of LLMs and generative AI in general.

First, we will discuss the concept and applications of generative AI in detail. Then, we will discuss the key technologies and frameworks that are used to build LLMs in this paper, and the RAG areas.

2.1.2 Generative AI trends

In recent times, as of 2023, the advancements in ultra-large AI technology have been rapid and significant. Table 2 illustrates the recent landscape of generative AI model releases. OpenAI unveiled GPT-4 in March, and Google introduced PaLM2 in May. Despite reducing the number of parameters compared to their previous models, these new iterations have been trained on approximately five times more tokens (text data), resulting in improved real-world performance. Moreover, prominent companies such as Samsung Electronics are in the process of developing or considering their own generative AI models.

Interest has also been growing in open-source LLM like Meta LLaMA and Falcon, which prioritize learning volume over model size. NVIDIA is pursuing the concept of an "AI Factory," which involves integrating AI models into corporate data centers. They are also developing small Large Language Models (sLLM) aimed at minimizing costs and providing easily deployable models within businesses.

In a notable instance, Naver Cloud introduced "HyperCLOVA X" a Korea-focused ultra-large AI model, in August. HyperCLOVA X is specialized for various fields such as e-commerce, finance, law, and education. It can be customized to a company's data and specific domain, and can be deployed using an API or Neuro Cloud approach. This progression indicates that the LLM market is transitioning from a focus on performance competition to a more specialized competition, marking a new phase in its evolution.

**Table 2** Status of Generative AI model releases in 2023

| Country | Company | Foundation Model | Parameters | Service | Source | Release Date |
|---|---|---|---|---|---|---|
| USA | OpenAI / MS | GPT-4 | Not disclosed | ChatGPT, MS Bing AI, MS Copilot, MS365 Copilot | Closed | 2023.03 |
| | Google | PaLM2 | 540B | Bard | Closed | 2023.05 |
| | META | LLaMA2 | 7~65B | - | Open | 2023.07 |
| | Stanford Univ. | Alpaca | 7B | LLaMA 7B Fine-tuning → LLaMA-7B | Open | 2023.03 |
| | Nomic | GPT4All v2 | 30~13B | LLaMA7B Fine-tuning Model | Open | 2023.04 |
| | Hugging Face | BLOOM | 176B | - | Open | 2022.07 |
| China | Huawei | PanGu 3.0 | 100B | Pangu-Weather, etc. | Open | 2023.07 |
| | Baidu | Ernie 3.5 | 130B | Ernie Bot 3.5 | Self-utilization | 2023.06 |
| | Alibaba | Tongyi Qianwen | 100B | DingTalk, Tmall Genie Open source service: ModelScope | Self-utilization Open | 2023.04 |
| South Korea | Naver | HyperClovaX | Not disclosed | Polaris Office AI, Lewis, etc. | Closed | 2023.08 |
| | LG | EXAONE2.0 | 300B | AI artist Tilda, etc. | Self-utilization | 2023.07 |
| | NC Soft | VARCO | Not disclosed | VARCO Art/Text/Human/Studio | Closed | 2023.08 |
| | KT | MI:DEUM2 | Not disclosed | GiGA Genie, AICC, AI care service | Self-utilization | Q3 2023 |
| | Kakao | Ko GPT2.0 | 6~65B | Providing specialized platforms for logistics, medical care, finance, etc. | Self-utilization | Q4 2023 |

In particular, open-source LLM have gained traction since the release of Alpaca, which is built upon the LLaMA architecture. Following Alpaca's introduction, various LLMs like GPT4All have continued to be unveiled. GPT4All, in particular, is based on the LLaMA 7B model and was inspired by Alpaca. It collected 800,000 prompt-response pairs from the GPT-3.5-Turbo model, encompassing code, conversations, and narratives. Among these pairs, around 430,000 were designed in an assistant-style prompt-response format, making them approximately 16 times larger in scale compared to Alpaca's dataset. A notable advantage of this model is its ability to run on CPUs without the need for GPUs.

The development of sLLM by startups is also progressing. LLM face hardware limitations and cost issues due to their extensive parameter count required for training. In contrast, sLLM aim to address these challenges by focusing on specific domains and languages, training on large datasets, and offering performance tailored to particular areas such as everyday conversations and domain-specific terminology. This development is highlighted as a potential solution to the limitations of traditional LLMs [4].

The introduction of various AI models from different companies is expanding the range of applications for generative AI. Generative AI is already being utilized in diverse fields like art, gaming, and entertainment. In recent times, the application of generative AI has extended to various industries. In the medical field, it can be used to identify medical conditions in patients and develop new drugs or treatment methods. In manufacturing, it can aid in creating new product designs and optimizing production processes. In the financial sector, generative AI can assist in developing new financial products and managing risks associated with financial transactions. This demonstrates how the utilization of generative AI is expanding across various industries, including healthcare, manufacturing, and finance.

## 2.2 Key Elements of Generative AI

### 2.2.1 Foundation Model

Generative AI models are categorized based on the type of output they produce, such as language models, image models, and video models. However, the current trend is the rapid development of multi-modal models that can simultaneously learn from both images and text, with their performance and capabilities evolving rapidly. These multi-modal models are also emerging as foundation models, becoming a cornerstone in AI technology. The concept of foundation models was first proposed by researchers at Stanford University, who published a paper titled "Opportunities and Risks of Foundation Models" Foundation models are trained using vast amounts of data, which includes text, images, audio, structured data, 3D signals, and more. These models are designed to perform tasks involving human creativity and reasoning. The term "Foundation Model" refers to a fundamental shift in the AI paradigm, highlighting their significance in the AI landscape [5].

In this context, extensive amounts of data are used for unsupervised learning to train the foundation model. Once trained, the model is distributed and can be fine-tuned or undergo in-context learning for downstream tasks according to the user's requirements. Foundation models can handle various types of data, allowing them to process any format of input, not limited to a single output type.

Foundation models exhibit the characteristic of emergence. This refers to the model's ability to derive knowledge for solving problems without being explicitly programmed beforehand. In the case of AI neural networks, the model can make decisions or infer the next steps autonomously based solely on the available data. This concept is rooted in the fundamental principle of emergence. As the volume of data continues to grow, the characteristic of emergence will play an even more significant role [6].

Furthermore, foundation models possess the characteristic of homogenization. This means that as the model gradually extracts generalized knowledge, having a consistent dataset becomes pivotal. With this, a massive foundation model can potentially tackle a wide range of problems. In essence, the homogenization feature empowers the model to derive comprehensive insights and knowledge from a unified dataset, enabling it to address various challenges [5].

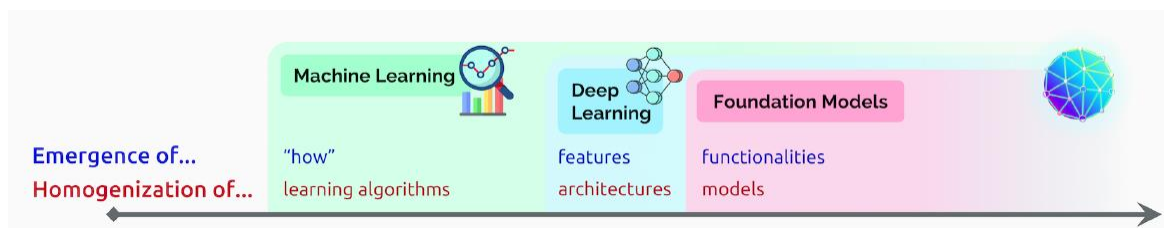

**Fig 2** The story of AI has been one of increasing emergence and homogenization

2.2.2 LLM (Large Language Model)

Generative AI models, particularly LLM, are widely used in chatbot services like ChatGPT. These models are trained on language data, such as text, and provide results through generating text-based responses. LLMs, such as OpenAI's GPT series and Google's BERT, have gained significant attention in the field of artificial intelligence. They are based on machine learning and deep learning technologies, requiring large-scale datasets and substantial computational power for training [7].

LLMs serve as foundational models for Natural Language Processing (NLP) and Natural Language Generation (NLG) tasks. To cope with the complexity and interconnectedness of language, these models undergo pre-training on vast amounts of data, followed by fine-tuning and techniques like in-context learning, zero/one/few-shot learning [8].

Notably, the quality of pre-training data strongly influences the performance of LLMs. Compared to smaller language models, LLMs have a higher demand for high-quality data for pre-training. The model's capacity relies heavily on the collection of training corpora and the methods used for pre-training. The following details describe the stages of training for LLMs, including data sources and preprocessing methods.

• **LLM Training Stages**

1) Data gathering and Preprocessing: The first step involves gathering the training dataset, which serves as the learning resource for the LLM. Data can be sourced from various places, including books, websites, articles, publicly available datasets, and more. To develop a capable LLM, pre-trained text datasets are used. The sources of pre-trained corpora can be broadly classified into two types: general data and domain-specific data. General data, like web pages, books, and conversational texts, are large and diverse. Due to their accessible nature, they are commonly utilized in most LLMs to enhance language modeling and generalization capabilities. Furthermore, there are studies that extend LLM training with more specialized datasets, such as multilingual data, scientific data, and code, to equip LLMs with specific task-solving abilities. Fig 3 depicts the typical process of data collection and pre-training for LLMs [9].

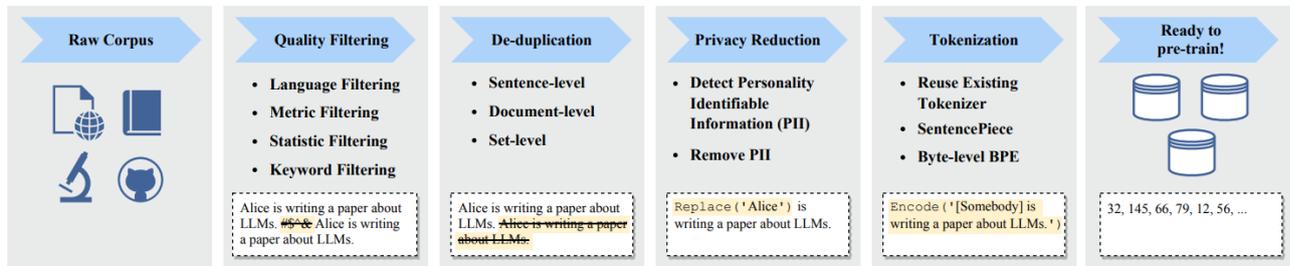

**Fig 3**. LLM Pre-Training Process

Popular public sources for finding datasets include platforms like Kaggle, Google Dataset Search, and Wikipedia's database. The next step involves cleaning and preparing the data for training. This can include tasks such as converting the dataset to lowercase, removing stopwords, and tokenizing the text into token sequences, which form the basic units for processing and analysis.

2) Model Selection and Configuration: Large models such as Google's BERT and OpenAI's GPT-3.5 commonly employ the Transformer deep learning architecture, which has been a prevailing choice for sophisticated NLP applications in recent years. As part of configuring the Transformer neural network, elements like the number of layers in the Transformer block, the number of attention heads, the choice of loss function, and hyperparameters must be specified. Model configuration can vary based on the desired use case and training data, and it significantly influences the training time of the model.

3) Model Training: The model undergoes supervised learning using preprocessed text data. During training, a sequence of words is presented to the model, and it learns to predict the next word in the sequence. The model adjusts its weights based on the difference between its predictions and the actual next word. This process is repeated millions of times until the model achieves a satisfactory level of performance. Due to the size of the model and the data, training a model requires significant computational power, often utilizing techniques like model parallelism to reduce training time. Training a large-scale language model from scratch demands substantial investment. Therefore, a more economical approach is fine-tuning an existing language model for specific use cases.

4) Evaluation and Fine-tuning: After training, the model's performance is evaluated using a separate test dataset that wasn't used during training. Based on the evaluation results, fine-tuning might be necessary to improve the model's performance. This could involve adjusting hyperparameters, changing the architecture, or further training on additional data. The goal is to enhance the model's capabilities based on the insights gained from evaluation.

2.2.3 Prompt Engineering

In ChatGPT, the quality of the generated responses greatly depends on how detailed and specific the prompts (questions or requests) are conveyed. This is why the exploration of prompt engineering, which involves finding combinations of prompt input values from LLM, plays a crucial role. Prompt engineering is important because it seeks to improve the quality of answers without the need for extensive parameter updates through large-scale data or fine-tuning processes. One effective approach to enhance answer quality is by providing example answer instances when prompting questions. This method guides the model to generate responses that are similar to the provided examples, thereby improving response quality. This is achieved without necessarily relying on massive data or fine-tuning procedures for parameter updates [2]. The approach of using examples within prompt instructions can be categorized into three types: Zero-Shot, where no examples are included; One-shot, where a single example is provided; and Few-shot Learning, where two or more examples are included. Providing a variety of examples in the prompt tends to yield better responses above a certain threshold [2]. Table 3. shows examples of each prompt method [10].

**Table 3** Zero/One-shot/Few-shot Learning

| Learning Type | Description |
|---|---|
| | **Prompt : Write a short alliterative sentence about a curious cat exploring a garden** |
| Zero-shot learning | [Example including prompt] - <br>[ChatGPT answer] A cat looks at flowers in the garden |
| One-shot learning | [Example including prompt] Peter Piper picked a peck of pickled peppers. <br>[ChatGPT answer] Curious cat cautiously checking colorful cabbages. |
| Few-shot learning | [Example including prompt] <br>Example 1: Peter Piper picked a peck of pickled peppers. <br>Example 2: She sells seashells by the seashore. <br>Example 3: How can a clam cram in a clean cream can? <br>[ChatGPT answer] <br>Curious cat crept cautiously, contemplating captivating, colorful carnations |

Furthermore, based on structure, functionality, and complexity, they can be classified into seven types as depicted in Fig 4.[11] Prefix prompts represents the simplest form of prompt wherein words or phrases indicating response type, format, and tone for control and relevance are added at the beginning. Cloze prompts is based on the idea of filling in blanks by generating masked tokens within the input text and requesting the language model to predict the missing words or phrases. On the other hand, an Anticipatory Prompt guides the conversation by anticipating the subsequent questions or commands based on experience or knowledge.

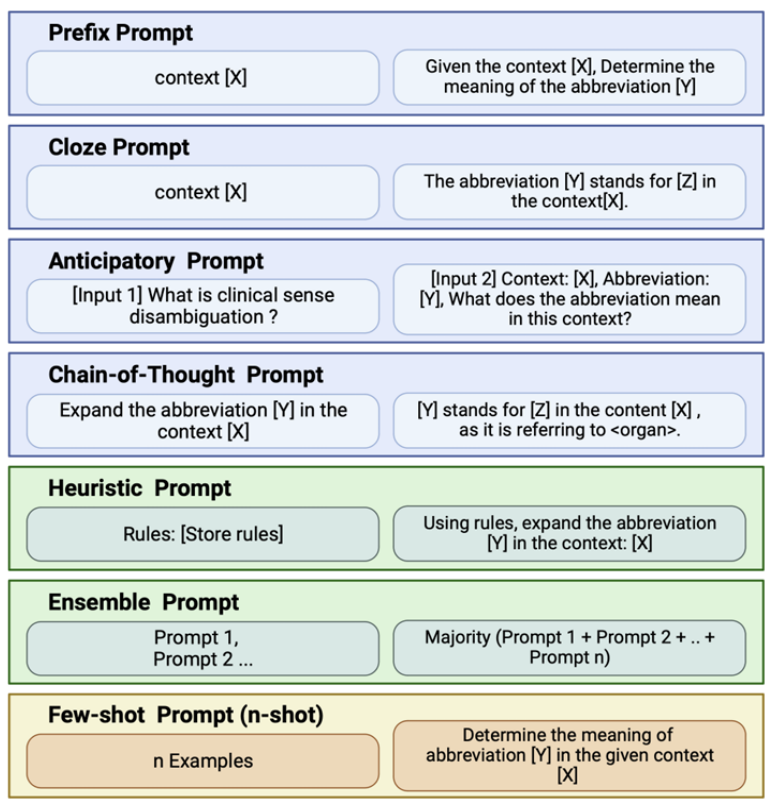

**Fig 4.** Types of Prompts : [X]: context , [Y]: Abbreviation, [Z]: Expanded Form

As the size of LLM models increases, they exhibit emergent phenomena that even developers cannot predict. One approach to finding accurate answers is by utilizing the Chain of Thought (CoT) prompts, where prompts guide LLMs to construct answers step by step, connecting information to generate accurate responses. This method of prompt engineering helps obtain precise answers for complex questions. Heuristic prompts is a rule-based prompt that involves breaking down complex queries into smaller parts to obtain comprehensive answers. On the other hand, the Ensemble prompts combines multiple prompts using a majority vote on aggregated outputs, aiming for a more comprehensive response. Various types of prompts are employed to generate multiple outputs for the same input, and the most common output is chosen as the final answer through this prompt approach.

As such, Prompt engineering is a technique used with LLM to generate creative content. LLMs are AIs trained on extensive text and code datasets, enabling them to generate text, translate languages, create diverse types of creative content, and respond to questions. Additionally, prompt engineering emphasizes enhancing the quality of prompts provided to LLMs to boost their creative text generation capabilities. Prompts serve as instructions and guidelines for LLMs, varying from simple to complex, and can specify the format, content, and style of the text LLMs are expected to produce.

Prompt engineering focuses on improving the following aspects of prompts:

• Clarity: Prompts should be easy to understand and follow, using specific language and examples.

• Relevance: Prompts should be relevant to the text LLMs are expected to generate.

• Creativity: Prompts should provide diverse expressions and ideas to encourage LLMs to produce creative and original text.

By doing so, the creative text generation capability of LLMs can be significantly enhanced. These learning approaches showcase the diverse ways in which artificial intelligence models acquire new information and make predictions. Utilizing prompt engineering effectively can enhance the model's learning and inference abilities.

2.2.4  Understanding the Generative AI Technology Stack

In order to comprehend the composition of the generative AI technology stack, we turn our attention to Fig 5. This illustration provides insight into the current state of the technological stack, with a focus on generative AI. Notably, this framework facilitates the analysis of generative AI technologies, categorizing them into distinct sections, each accompanied by a selection of well-established vendor examples [12]. Within the emerging landscape of generative AI, enterprises are engaged in developing proprietary models, leveraging third-party generative AI through APIs, or constructing applications by adapting finely-tuned open-source models to meet their specific needs.

• Application Layer: In the Applications layer, users either run their own model pipelines ("E2E Apps") or utilize third-party APIs for generative AI models (e.g., Jasper and Copilot).

• Model Layer: This layer drives AI products provided through proprietary APIs or open-source checkpoints (requiring hosting solutions). The Foundation Models encompass both private-source proprietary models (e.g., GPT-4) and open-source models (e.g., Stable Diffusion), along with model hubs that share and host Foundation Models.

• Infrastructure Layer: This tier encompasses the platforms and hardware (e.g., cloud platforms and hardware) responsible for executing training and inference workloads for generative AI models.

• Orchestration and Monitoring Layer: Within the generative AI stack, similar to this one, this layer incorporates model monitoring capabilities to deploy, understand, and safeguard the distribution of these models.

As revealed in Fig 5, the generative AI technology stack illustrates the intricate interplay among these layers, offering insights into the dynamic landscape of generative AI technology and its evolving applications [12].

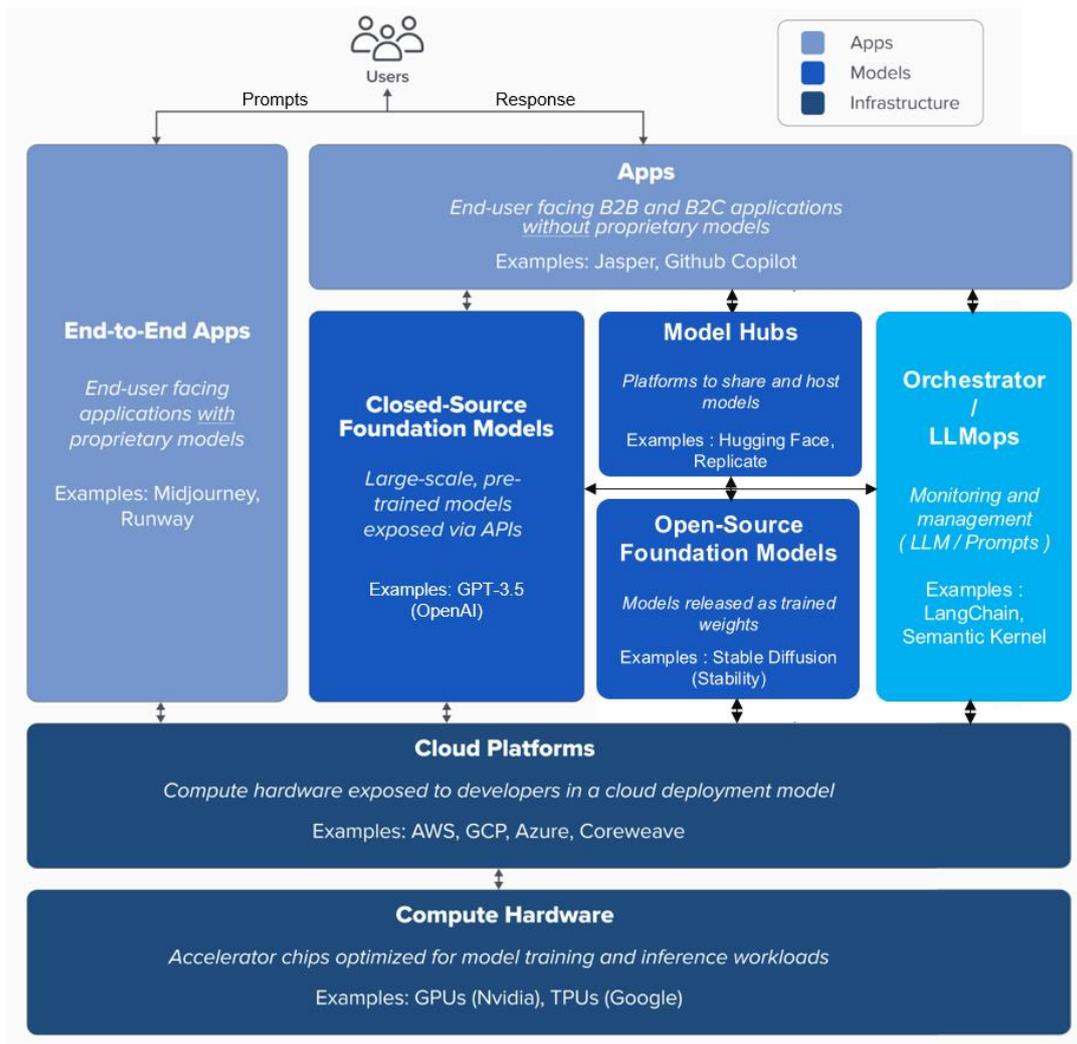

**Fig 5**. Generative AI Tech. Stack

## 2.3 Overview of the RAG Model

The advancement of conversational AI models utilizing LLM, such as ChatGPT, has garnered significant interest across various domains. Applications developed using LLMs are gaining attention, including scenarios like financial services where providing up-to-date information through customer chatbots is crucial. However, there are challenges related to limited answer capabilities due to insufficient information within LLMs. The phenomenon known as the "hallucination problem" arises when models generate stories and information that seem plausible but are actually fabricated, as the model creatively fills in gaps in its knowledge. This can lead to situations where seemingly plausible but incorrect information is generated.

To address these challenges, strategies have been proposed, such as fine-tuning LLMs with new data or directly injecting information into the prompt context. However, the former approach incurs significant costs, while the latter approach of embedding all information into prompts is not practical. As an alternative, the RAG (Retrieval-Augmented Generation) model has emerged. This model stores information in databases and retrieves the necessary information to provide it to the LLM when needed. By providing LLMs with relevant questions and associated reference materials beforehand, the model utilizes these references to generate more accurate and reliable answers.

In this section, we delve into the details of the RAG model, which is employed as an implementation method in this paper. The RAG model stores information, leverages databases, and enhances the generation process through the inclusion of relevant reference materials, ultimately leading to improved answer quality and reliability.

2.3.1 Architecture of the RAG Model

The RAG (Retrieval-Augmented Generation) model is designed for text generation tasks, performing a process that involves retrieving information from given source data and utilizing that information to generate desired text. Fig 6 illustrates the data processing pipeline for RAG usage, involving breaking down the original data into smaller chunks and converting text data into numerical vectors through embedding, which are then stored in a vector repository [13].

• Source Data Collection and Preparation: Relevant source data is required for the model's training and utilization. This data can include documents, web pages, news articles, etc. It forms the foundation for the model to search for and generate content.

• Chunking of Searchable Units: Source data is divided into smaller units known as chunks. Chunks are typically small text fragments, such as sentences or paragraphs, making it easier to search for and utilize information at this granular level.

• Embedding: Generated chunks undergo embedding, a process of converting text into meaningful vector representations. Pre-trained language models are often used to transform text into dense vectors, capturing the meaning and related information in vector form.

• Construction of Vector Database: A vector database is built based on the embedded chunks. This database represents the positions of each chunk within the vector space, enabling efficient retrieval and similarity calculations.

• Search and Information Integration: To retrieve information relevant to the context of the text to be generated, appropriate chunks are searched within the vector database. The retrieved chunks are decoded back into original text data to extract information, which is then utilized during the generation process.

• Text Generation: Using the retrieved information as a basis, text is generated. Users can specify the type, length, and linguistic style of the text to be generated. The RAG model is designed to seamlessly integrate information retrieval and generation processes, aiming to produce more accurate and meaningful text outputs.

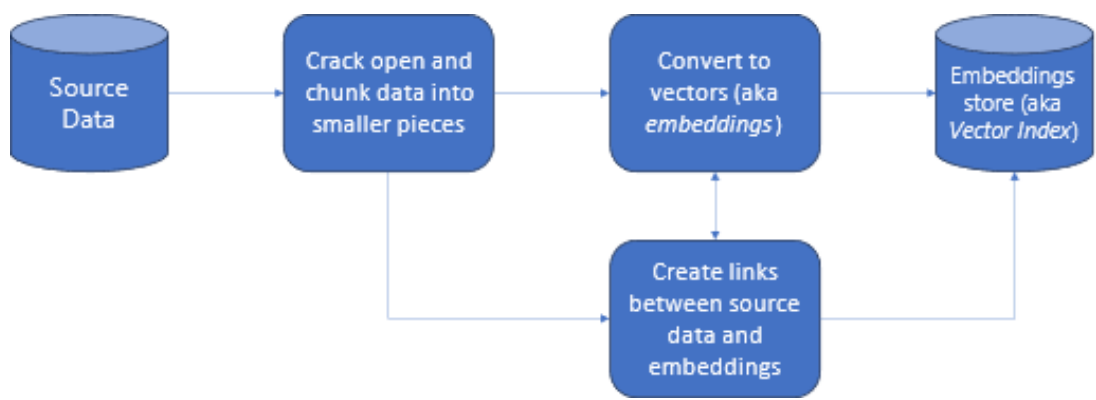

**Fig 6** RAG Model Diagram (Microsoft, 2023)

Indeed, the RAG model stands as an innovative methodology that combines retrieval and generation to enable more accurate and meaningful text creation. This architecture merges the strengths of information retrieval with the creativity of text generation, making it applicable across various domains and applications.

2.3.2  Using Embedding in LLM

In the context of LLM (Language Model), embedding, or embedding vectors, refer to the representation of text as fixed-size arrays of floating-point numbers. These arrays form real-valued vector shapes. When a specific word, sentence, or document is input into an embedding generation model, it outputs a vector composed of floating-point values. Although these vectors might be hard for humans to directly comprehend, calculating the distances between embeddings of different words or documents can reveal semantic relationships.

In recent times, it has become common to train language models like LLMs with neural network structures using large-scale document collections to create "Learned Embeddings." During this training process, the model takes in various words as input and learns semantic relationships by making embedding vectors of contextually similar words closer together and those of dissimilar words farther apart. This approach allows the model to capture meaningful relationships within the context.

Embedding can utilize the following two methods to identify semantic relationships between different words or documents.

- **Semantic Search**: Used when multiple documents exist and one of them must be searched based on meaning or compared to each other.

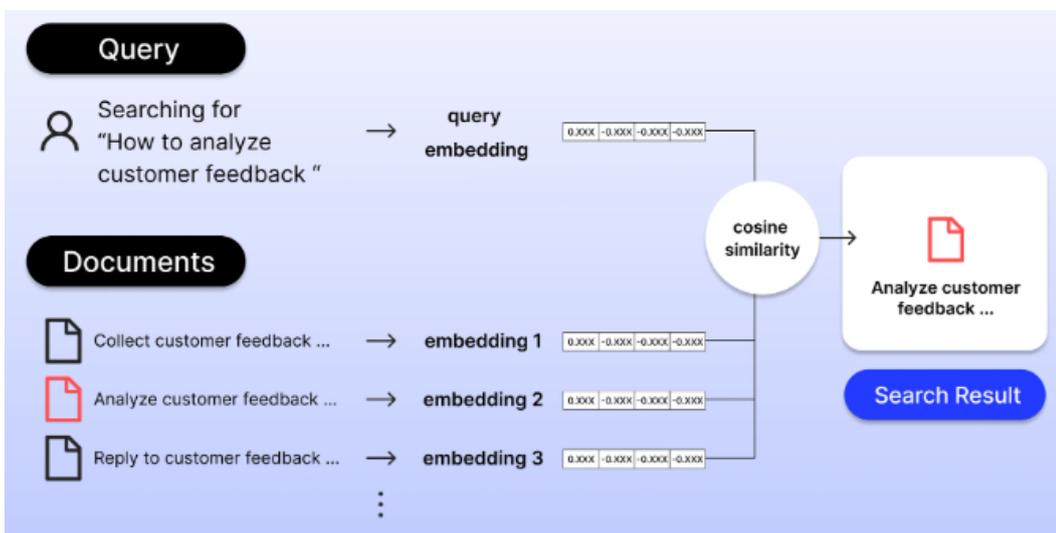

**Fig 7** 'Semantic Search' using Embedding

Semantic Search is a function that finds and presents documents that are semantically related to the text query presented by the user. The Semantic Search process using embedding is shown in Fig 7 [3].
1) Calculate embeddings for each document in a collection and store them in a repository (e.g., local drive or vector database).
2) Compute the embedding for the query text.
3) Calculate the cosine similarity between the query embedding and each document embedding, sorting the documents based on similarity.
4) Retrieve and return the text of the top k documents from the sorted results.

- **Question Answering**: Question Answering is a method to provide additional information to LLM for generating results. While LLM possesses general knowledge of publicly available internet information, it lacks information about internal corporate or personal private data. Therefore, to obtain answers to questions related to private information, the text containing that information needs to be included in the prompt to LLM. However, current LLM services have limitations on the length of input text, which requires breaking down long texts into smaller chunks. In cases where longer texts need to be processed, relevant chunks are selected and included along with the question in the prompt. This approach is a crucial consideration when implementing Question Answering functionality.

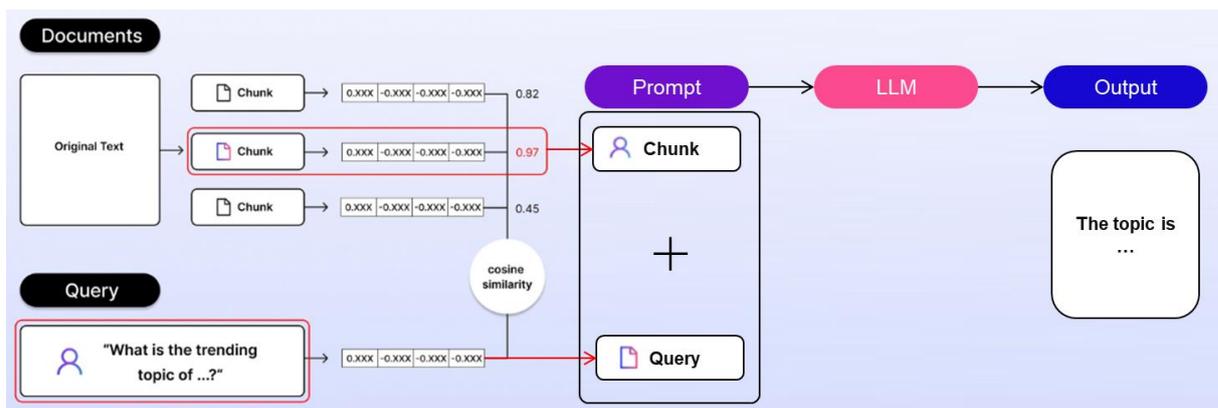

**Fig 8** 'Question Answering' using Embedding

The Question Answering process using general embedding, which allows LLM to answer a given question based on information additionally injected, is shown in Fig 8 [3].

1) Divide the entire information-rich text into smaller chunks and calculate embeddings for each chunk.
2) Compute the embedding for the query.
3) Calculate cosine similarity between query embedding and each chunk embedding, sorting the chunks based on similarity.
4) Retrieve the text of the top k chunks, which is then added to the prompt - k is determined by the maximum text length allowed by the LLM service.
5) Input the completed prompt into the LLM and return the generated answer.

APIs like OpenAI's Embedding API are commonly used to extract effective embeddings from text. These approaches enhance the capability of LLMs to understand and respond to queries that involve searching for information or generating answers based on specific context.

2.3.3 Vector Database

A vector database is a novel type of database developed to address the problem of long-term memory deficiency in LLMs. This database is specialized in efficiently storing and managing high-dimensional real-valued vector indices. Unlike traditional databases, a vector database represents queries in the form of real-valued vectors (embeddings) and supports a method to extract data with similar vectors. Therefore, it is optimized for storing and utilizing vectors obtained as results from AI models. The general pipeline for a vector database is illustrated in Fig 9 [14].

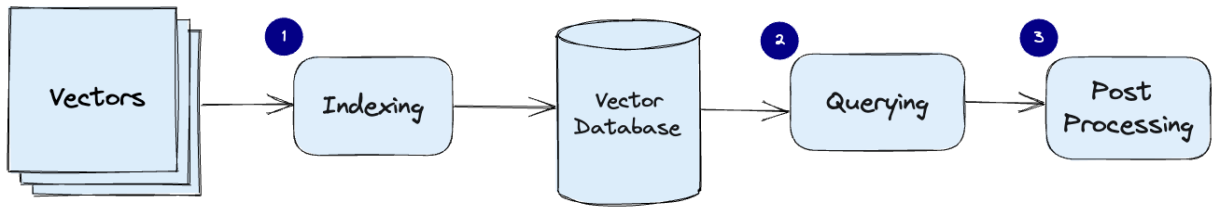

**Fig 9** Pipeline for a vector database

1) Indexing: A vector database employs algorithms such as Product Quantization (PQ), Locality-Sensitive Hashing (LSH), or Hierarchical Navigable Small World (HNSW) to index vectors. This step maps vectors to data structures that enable faster searches.
2) Querying: In the vector database, indexed query vectors are compared to indexed vectors in the dataset to find the nearest neighbors. The similarity metric used in the indexing is applied during this process.
3) Post Processing: Depending on the scenario, the vector database may retrieve the nearest final approximate data from the dataset and perform post-processing to return the ultimate result. This step might involve re-ranking the nearest neighbors using different similarity measurements.

As Claypot AI founder Chip Huyen mentioned, "If 2021 was the year of graph databases, then 2023 is the year of vector databases" highlighting the increasing interest in this field [15]. While standalone vector indexes like Chroma and FAISS (Facebook AI Similarity Search) enhance vector embedding search, they lack robust management functionalities. On the other hand, vector databases address limitations of standalone vector indexes, such as scalability issues, cumbersome integration processes, absence of real-time updates, and embedded security measures. Solutions like Pinecone offer stable cloud-based hosting, while Weaviate, Vespa, and Qdrant are open-source single-node databases. Chroma and FAISS serve as local vector management libraries and, while not strictly databases, they can easily complement other AI-related tools like LangChain and OpenAI embedding models for lightweight embedding search purposes.

2.3.4 Orchestration Framework for LLM service implementation

Two prominent orchestration frameworks for implementing LLM services are LangChain and Semantic Kernel. Both of these frameworks share the goal of efficiently utilizing the capabilities of LLM and providing users with a convenient user experience.

- **LangChain Framework**

    LangChain is an open-source framework that emerged in October 2022 and has garnered close to 44,000 stars on GitHub as of June 2023. This project has experienced explosive growth within its own community and is actively building an ecosystem on top of the framework. LangChain utilizes LLM to perform various natural language processing tasks. It supports a range of LLM models from sources like OpenAI and Hugging Face. Users have the flexibility to choose and utilize models that best suit their specific requirements within the LangChain framework. At the core of LangChain is the concept of chaining LLM prompts and external source executions (such as calculators, Google searches, sending Slack messages, or running source code) to perform a sequence of actions. Using LLM, LangChain can perform various tasks including translation, summarization, question answering, text generation, and natural language inference by chaining together different steps.

    - Translation Service: LangChain allows the deployment and management of translation models,

providing support for a variety of languages.

- Summarization Service: Summarization models can be deployed and managed within LangChain, offering summarization capabilities across different topics.

- Question Answering Service: LangChain enables the deployment and management of question answering models, catering to various domains.

- Text Generation Service: The framework facilitates the deployment and management of text generation models, supporting the creation of text in various formats.

- Natural Language Inference Service: LangChain can deploy and manage natural language inference models, facilitating diverse natural language inference tasks.

- **Semantic Kernel Framework**

Semantic Kernel is an open-source SDK developed by Microsoft, designed to facilitate the seamless integration of LLM into existing applications. It aims to provide developers with a user-friendly way to incorporate LLM capabilities into their applications. The SDK supports a variety of programming languages including C#, Python, Java, among others. It also enables integration with a range of LLM services such as Azure OpenAI, OpenAI, and Hugging Face.

Semantic Kernel leverages AI technologies such as natural language processing, machine learning, and planning to imbue applications with the following capabilities:

- Natural Language Interface: Enabling users to interact with applications using natural language.

- Text Generation: Creating diverse forms of creative textual content, including text, code, scripts, musical compositions, emails, and letters.

- Question Answering: Furnishing users with informative and comprehensive answers to their queries.

- Task Automation: Facilitating the automatic execution of tasks within applications as directed by users.

Indeed, both LangChain and Semantic Kernel stand as robust frameworks for implementing LLM services. Users have the flexibility to choose the most fitting framework based on their specific requirements. Whether opting for LangChain's chaining capabilities, which intertwine LLM prompt execution and external source invocation, or embracing Semantic Kernel's seamless integration of AI technologies for natural language interfaces, text generation, question answering, and task automation, these frameworks offer a versatile choice for users seeking to implement LLM-based services.

2.3.5 AI Chatbot

Through AI chatbots like ChatGPT, generative AI services can be harnessed, allowing users to input prompts for desired questions and receive corresponding responses. Today, AI-driven chatbot systems encompass not only simple question-answer scenarios like FAQs, but have advanced to the level of analyzing human emotions and intentions to provide tailored responses [16]. Recent advancements in Natural Language Understanding (NLU) technology, coupled with the utilization of context models and Transformer language models, have enabled the handling of intricate conversations. Furthermore, the maturation of Speech-to-Text (STT) and Text-to-Speech (TTS) technologies has popularized voice-based services [16].

Chatbots are AI-based interactive software that facilitate conversations between individuals and service bots, offering appropriate answers and relevant information through text or voice interactions [17]. Sánchez-Díaz, Ayala-Bastidas, Fonseca-Ortiz & Garrido define chatbots as intelligent agents that enable users to engage in conversations typically through text or voice [18, 19]. Chatbots leverage Natural Language Processing (NLP) technology to comprehend human queries and deliver responses in a conversational and natural manner [20].

NLP consists of two primary components: Natural Language Understanding (NLU), which reads and comprehends natural language questions, and Natural Language Generation (NLG), which generates responses in natural language [20]. NLG employs Machine Learning (ML) to learn from extensive corpora and generate answers. In 2020, OpenAI introduced the GPT-3 (Generative Pre-trained Transformer-3) NLP model, which was trained on a dataset of 300 billion tokens with 175 billion parameters. In 2021, Microsoft and NVIDIA unveiled the MT-NLG (Megatron-Turing Natural Language Generation) model, boasting a massive 530 billion parameters in a LLM. This advancement enabled automated conversation generation and translation capabilities [20].

Furthermore, in November 2022, OpenAI introduced the GPT-3.5 model, which was developed through a learning process involving reinforcement learning from human feedback (RLHF). This model, an AI chatbot named ChatGPT, was trained to generate human-like natural language in interactive conversations. Within just two months of its release, ChatGPT garnered immense attention, surpassing over 100 million monthly users [20]. During the initial stages of chatbot adoption, due to concerns over reliability, chatbots were primarily employed for simple tasks such as answering basic queries in customer support functions. However, more recently, chatbots are being directly integrated into business processes, often in conjunction with other solutions like Robotic Process Automation (RPA) and Optical Character Recognition (OCR), to enhance efficiency [19]. Additionally, chatbots are being leveraged in collaboration with automation platforms to serve as channels for status updates, task notifications, and result dissemination [20]. In particular, in open-domain chatbot dialogues like those with ChatGPT, the interaction interface occurs through prompts to ChatGPT and LLM models. When applying parameters for interfacing between applications, it's crucial to be cautious about the information involved, especially with regard to personal data. While laws exist to prevent developers from collecting and using user data without consent, in practice, users often find it difficult to discern how much data developers are collecting and where it's being stored [21].

## 2.4 Summary of Generative AI Literature Review

The existing literature on generative AI and related technologies has provided insights into various concepts and techniques. However, there is a lack of comprehensive guidance on how to effectively combine and manage these diverse technologies to apply them as LLM services in real-world business scenarios. Many resources focus on introducing generative AI and LLM or emphasize security concerns related to utilizing generative AI with publicly available information. Moreover, there is limited research that systematically addresses the utilization of internal business information to resolve security issues within a framework that effectively orchestrates various generative AI technologies.

To overcome these limitations, this study aims to address the shortcomings in existing generative AI literature and its practical application in business contexts. By considering the gaps and deficiencies, a framework for implementing LLM services will be employed to systematically present a method for integrating and applying generative AI technologies. The ultimate goal is to provide a comprehensive approach that effectively utilizes generative AI techniques within a framework, addressing both security concerns and practical implementation for real-world business scenarios.

## III. Methods

In this chapter, we outline a comprehensive framework for implementing generative AI services by effectively combining and orchestrating various technologies within the previously discussed generative AI technology stack. We will describe how each component of the framework is applied to achieve efficient and harmonious integration of the diverse generative AI technologies. The following sections detail the implementation process for each technology component within the framework:

3.1 Framework for Implementing Generative AI Services using RAG Model

Based on previous research, we have designed a comprehensive framework for implementing generative AI services using the Retrieval-Augmented Generation (RAG) model. The framework encompasses a series of procedures and core functionalities, as depicted in Fig 10. The diagram conceptually illustrates the process of utilizing LLMs to retrieve information from documents and outlines the key steps involved. The major components of the framework are detailed below.

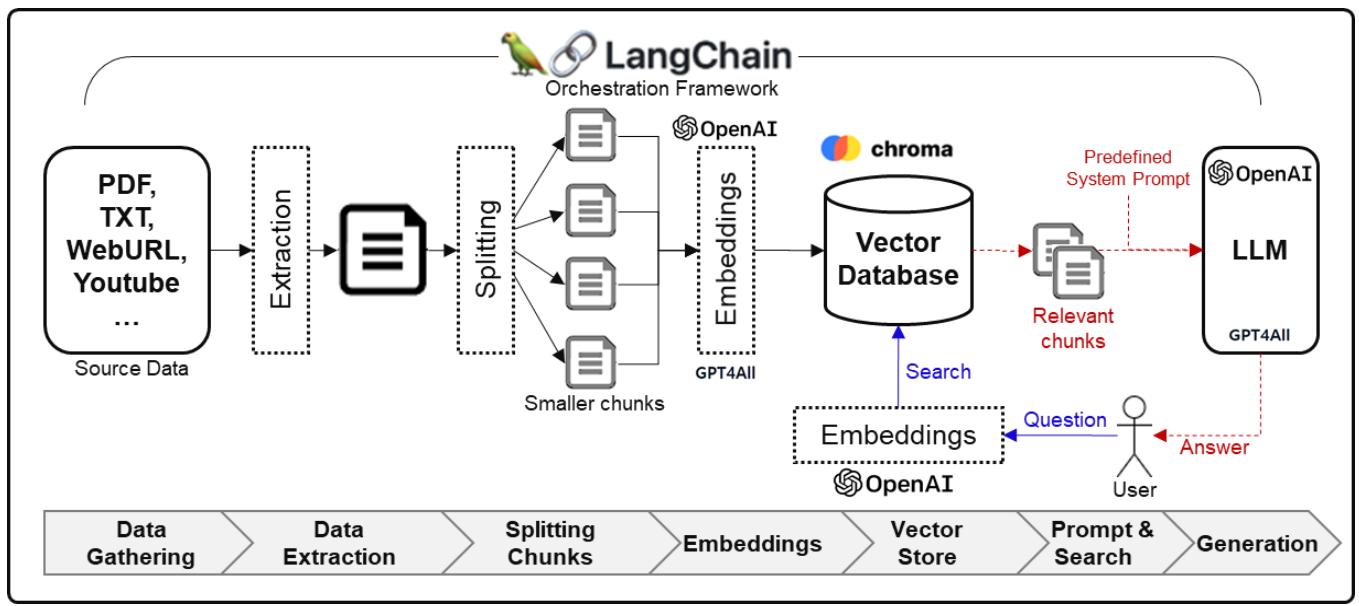

**Fig 10** Framework for Implementing Generative AI Services using RAG Model

3.2 RAG model and LangChain integration implementation process

Within the framework for implementing generative AI services, various solutions exist for each step of the process. Considering the awareness and cost aspects of these solutions, this study has proposed a framework that primarily relies on open-source products in alignment with the findings from Chapter 2. In this context, the framework takes advantage of both OpenAI's proprietary models and open-source models to trigger the generative AI capabilities of LLMs. The composition of the framework is illustrated in Fig 10.

For the overall orchestration framework, LangChain is adopted, while specific tasks like chunking and embedding are accomplished using a combination of OpenAI models and GPT4All. The vector repository is facilitated through Chroma DB, chosen for its ease of implementation. Regarding the LLM component, OpenAI's GPT-3.5-turbo model and GPT4All are integrated, allowing developers to harness diverse choices to achieve optimal development outcomes.

### 3.2.1 RAG based implementation procedure

The RAG model, as discussed in the architecture of the RAG model in Chapter 2, is a search-augmented generative model used to retrieve and generate responses based on information relevant to given questions or topics. Each step follows the procedure outlined in Fig 9.

1) Source Data Collection and Extraction: During the data collection phase, both structured and unstructured data are gathered. Structured data is stored in standardized formats such as CSV, JSON, or XML, while unstructured data is stored in formats like PDF, TXT, HTML, images, and videos. Preparatory materials related to the task, such as regulations, user manuals, and terms and conditions, are loaded into LangChain using the LangChain module.

2) Chunk Generation: Source data is processed to split it into smaller units known as chunks. These chunks typically consist of sentences or paragraphs, serving as smaller text fragments that can be used to search and retrieve information from LLM. LangChain's module is utilized to split data into chunks that are suitable for retrieval.

3) Embedding: The generated chunk-level text data is transformed into numerical vector representations. This step involves mapping words or sentences to vectors, and libraries provided by OpenAI or GPT4All can be employed for this purpose.

4) Building the Vector Database: Based on the embedded chunks, the vector database is constructed. This database represents the positions of each chunk in the vector space, facilitating efficient search and similarity calculations. Typically, content from each document is included, embeddings and documents are stored in the vector repository, and documents are indexed using embeddings. Tools like Chroma or FAISS for vector indexing can be used.

5) Integration of Prompt and Search Results: This step involves searching for information based on the prompted question and integrating relevant information. To search for contextually relevant information based on the prompt, appropriate chunks are retrieved from the vector database. These retrieved chunks are then sent to LLM to aid in the response generation process. Various search engines available within LangChain for vector store similarity search are utilized.

6) Answer Generation: Using the retrieved information as a basis, the response text is generated. At this stage, the type, length, and linguistic style of the generated text can be specified. LLM, such as OpenAI's GPT-3.5-turbo model or GPT4All, uses the similarity search module in LangChain to retrieve relevant documents and generate responses.

The RAG-based implementation procedure outlined above illustrates how the RAG model, in combination with LangChain, can be effectively integrated into the generative AI service framework. This comprehensive approach covers data collection, processing, embedding, search, and response generation, ensuring that the service provides accurate and contextually relevant answers to user queries.

## IV. Experiment

In this chapter, the generative AI service implementation framework introduced in Chapter 3 is utilized to implement various scenarios based on enterprise internal data using the integrated RAG model and LangChain according to the implementation procedure. This provides a series of practical examples for each implementation step. Through these examples, we explore the methods of implementation and consider the factors to be taken into account during the construction process.

## 4.1 Development Environment

The solutions and development framework applied in this case follow the basic approach presented in Fig 9. The aim is to provide an implementation method using LangChain and the OpenAI API to transform documents. The process includes dividing the documents into chunks, converting them into embeddings, and storing them in ChromaDB. This approach ensures effective access to the information required to provide context-based answers. The implementation is carried out using the Python programming language. Python is widely used in AI development due to its diverse libraries and frameworks that are well-suited for AI development.

The development environments for each implementation component are as follows:
- Orchestration Framework: LangChain
- Data Extraction and Chunking: LangChain Module
- Embeddings: OpenAI, GPT4All
- Vector Database: Chroma, FAISS
- LLM: OpenAI GPT-3.5-turbo Model, GPT4ALL
- Python Development Environment: Google Colab

## 4.2 Implementation Results by Step

### 4.2.1 Installing Basic Libraries and Setting OpenAI API Key

Initially, install the essential libraries including LangChain, which serves as the overarching orchestration framework, and the OpenAI library. For OpenAI, unlike open-source libraries, you need to obtain an API key and conFig it for usage during implementation. It is recommended to manage the key securely, for instance, by storing it in a .env file in a specific location. In your development source code, read the content of the file containing the key to authenticate the API calls. Fig 11 demonstrates the installation of the dotenv library for managing environment variables and successfully reading the .env file containing the key.

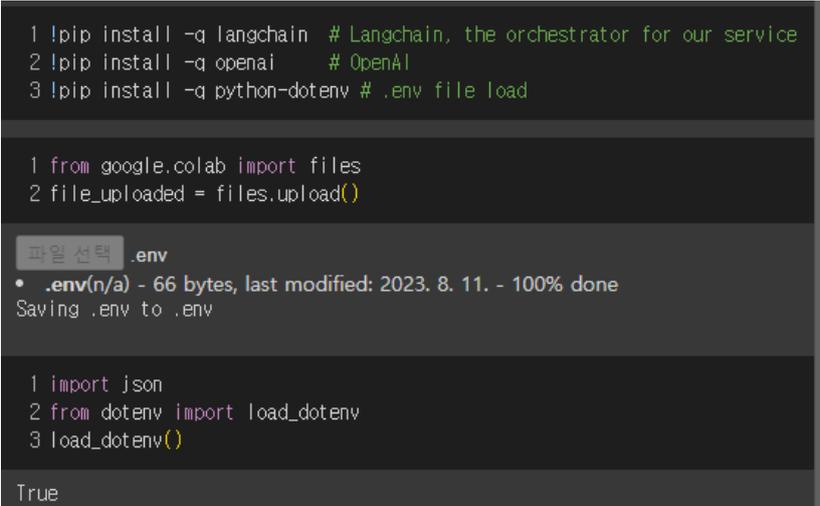

Fig 11. Basic library installation and OpenAI API key setting

### 4.2.2 Implementation of Source Data Gathering and Data Extraction Step

Source data can exist in various forms, including unstructured data within documents like docx, xlsx, csv, pptx from the MS Office suite, or other formats like txt, pdf. Additionally, data might be available in the form of web pages or videos from platforms like YouTube. To extract information from such diverse and unstructured raw data, loading the data is the first step. Table 4 outlines the implementation code for loading different types of documents, specifically MS Office documents like 'Leave and Return Management Standards.docx' as well as unstructured documents like 'Dress Code Standards.pdf' and 'Payment Insurance Calculation.txt' which are stored in designated locations.

**Table 4** Source data load type

| Data Type | Python Code |
|---|---|
| PDF | from LangChain.document_loaders import PyPDFLoader<br>loader = PyPDFLoader("/content/Dress Code Standards.pdf ") |
| TXT | from LangChain.document_loaders import TextLoader<br>loader = TextLoader("//content/'Payment Insurance Calculation.txt") |
| DOC | from LangChain.document_loaders.word_document import UnstructuredWordDocumentLoader<br>loader = UnstructuredWordDocumentLoader("Leave and Return Management Standards.docx") |

Additionally, Fig 12. demonstrates the implemented code for loading PDF documents, which is a common type of unstructured document. The remaining types of documents have also been implemented for loading similar to Table 4.

```
1 import os
2 os.listdir() # source data file list

['.config',
 '.env',
 'Payment Insurance Calculation.txt',
 'Dress Code Standards.pdf',
 'Leave and Return Management Standards.docx',
 'sample_data']

1 !pip install -q pypdf  # Library to extract text from pdf document
2 from langchain.document_loaders import PyPDFLoader
3 loader = PyPDFLoader("/content/Dress Code Standards.pdf")
4 pages = loader.load_and_split()
5 data = loader.load()
6 print(f"{len(data)} documents, contain {len(data[0].page_content)} words")

2 documents, contain 3178 words
```

**Fig 12** Source Data gathering and Data load

### 4.2.3 Implementing the Splitting Chunks Step

The next step involves using the LangChain integration feature again to split the large source document into smaller chunk units for embedding and vector storage. Fig 13. showcases the code implementation to split documents into chunks, where each chunk contains a maximum of 500 characters ('chunk_size = 500').

```
1 from langchain.text_splitter import RecursiveCharacterTextSplitter
2
3 text_splitter = RecursiveCharacterTextSplitter(chunk_size = 500, chunk_overlap = 0)
4 all_splits = text_splitter.split_documents(data)
```

**Fig 13** Splitting Chunks

### 4.2.4 Implementation of Embeddings and Vector Database Construction Steps

First, Chroma, the vector database used to store embeddings, is installed. The chunks are then sent to the OpenAI embedding engine to be transformed into numerical vectors. Each of these vectors, generated in this manner, is stored in Chroma for future retrieval, as illustrated in Fig 14. Chroma, which is open source, functions as an in-memory store that does not retain content when sessions are restarted.

```
1 !pip install -q chromadb

1 from langchain.embeddings import OpenAIEmbeddings
2 from langchain.vectorstores import Chroma
3
4 vectorstore = Chroma.from_documents(documents=all_splits, embedding=OpenAIEmbeddings())
```

**Fig 14** Embedding

4.2.5 Implementation of question (Prompt) and search result integration steps

In this step, we utilize similarity search to retrieve relevant chunks for each query. When a user submits a query, the first task is to convert it into an embedding to find the most relevant chunks within the Chroma DB. The code to achieve this using LangChain is presented in Fig 15. For instance, if a user asks, "What is the work dress code for male employees?" the vector database containing the 'dress code standards.pdf' document will be queried to find chunks with content similar to the query. Furthermore, the implementation also suggests similar queries to enhance the user experience.

```
1 question = "What is the work dress code for male employees?"
2 docs = vectorstore.similarity_search(question)
3 print(f"Question length : {len(docs)}, document : {len(data)}, words :  {len(data[0].page_content)}")

Question length : 4, document : 2, words :  3178

1 # Similar questions recommended
2 import logging
3 from langchain.chat_models import ChatOpenAI
4 from langchain.retrievers.multi_query import MultiQueryRetriever
5
6 logging.basicConfig()
7 logging.getLogger('langchain.retrievers.multi_query').setLevel(logging.INFO)
8
9 retriever_from_llm = MultiQueryRetriever.from_llm(retriever=vectorstore.as_retriever(),
10                                                   llm=ChatOpenAI(temperature=0))
11 unique_docs = retriever_from_llm.get_relevant_documents(query=question)
12 len(unique_docs)

['1. What are the dress code requirements for male employees at work?', '2. Can you provide information on
```

**Fig 15** Question and search results

4.2.6 Implementation of answer Generation step

In the "Answer Generation" step, after retrieving the four closest chunks from the Chroma DB based on semantic similarity to the user query, these chunks are provided to the LLM to generate a response. These chunks serve as context to enable the LLM to produce a coherent and contextually relevant answer to the user's question. The implementation ensures that the LLM has the necessary information to craft an appropriate response.

```
1 from langchain.chains import RetrievalQA
2 from langchain.chat_models import ChatOpenAI
3
4 llm = ChatOpenAI(model_name="gpt-3.5-turbo", temperature=0)
5 qa_chain = RetrievalQA.from_chain_type(llm,retriever=vectorstore.as_retriever())
6 qa_chain({"query": question})
```

{'query': 'What is the work dress code for male employees?',
 'result': 'The work dress code for male employees is business casual attire. They should wear a suit jacket, primary color cotton pants, formal or neat casual shoes, a belt, and socks that match their clothes. They should avoid wearing sandals, boots, walkers, slippers, excessive accessories, and hats.'}

```
1 question = "What is the purpose of the dress code?"
2 qa_chain({"query": question})
```

{'query': 'What is the purpose of the dress code?',
 'result': 'The purpose of the dress code is to ensure that employees maintain professionalism and decorum in their attire while reflecting the individuality of each employee.'}

```
1 question = "What dress code should male employees avoid?"
2 qa_chain({"query": question})
```

{'query': 'What dress code should male employees avoid?',
 'result': 'Male employees should avoid wearing revealing or tight clothing (such as sleeveless or tank tops), t-shirts with sexually or politically aggressive content, excessively short or tight skirts, ripped jeans or jean skirts, primary color cotton pants, slippers, long boots, sandals without back straps, and sports sandals. They should also avoid wearing jackets, as suit jackets are not appropriate for the dress code.'}

**Fig 16** Answer generation using LLM

Fig 16 demonstrates the provided answer for the question "What is the work dress code for male employees?" Additionally, when further questions like "What is the purpose of the dress code?" and "What dress code should male employees avoid?" are posed, Fig 17 illustrates how the content from the document 'Dress Code Standards.pdf' is effectively retrieved from the vector database and used to generate accurate responses. This showcases the successful operation of the system in generating contextually relevant answers based on the user's queries.

### Dress Code Guidelines for Work

**1. Purpose**
Taking into account the individuality of each employee according to the nature of the business, employees are expected to wear work attire that reflects professionalism and decorum.

**2. Considerations for Dress Selection**
(1) Maintain honor and dignity as professionals.
(2) Always uphold neat and clean attire.
(3) Choose attire based on multiple objective criteria.
(4) Wear clothing that does not offend others or clients.

**3. Applicability**
(1) All employees.
(2) For employees working with client companies, follow the dress code of the respective client.

**4. Dress Standards**
(1) Maintain a neat and simple business casual attire, with the option of jeans and sneakers.
(2) Avoid attire that is excessively revealing or flashy.
(3) Shorts are permissible only from June to September.
  - Considering professional etiquette, shorts should be of 4 to 5 inches above the knee length, and should have a modest design, either formal or made of cotton fabric.
  - Extremely short shorts (revealing the upper thigh while standing), athletic wear, overly flashy patterns, etc., should be avoided.
  - Footwear should not include slippers; loafers, sneakers, or other well-maintained shoes that suit shorts are appropriate.

□ **Attire for Male Employees**

| Category | Business casual | Clothing to Avoid |
|---|---|---|
| Jacket | · Suit jacket<br>※ Suit jacket can be removed in hot weather | · Denim (Jean) jacket, parka, sportswear |
| Shirt | · Dress shirt (short-sleeved shirt available)<br>· Casual shirts with collars, T-shirts<br>· Neat knit without collar is possible<br>※ No-Tie allowed | · Hoodie, sleeveless, round tee, etc.<br>· Clothes that are too revealing<br>· T-shirts printed with sexually/politically aggressive content, etc. |
| Pants | · Suit pants<br>· Neat cotton pants<br>※ Shorts available in summer | · Ripped jeans, cargo pants, etc.<br>· Primary color cotton pants |
| Shoes | · Formal shoes<br>· Neat casual shoes | · Sandals, boots, walkers, slippers |
| etc | · Wear a belt<br>· Socks (color that matches clothes) | · Excessive accessories, hats |

□ **Attire for Female Employees**

**Fig 17** Part of the contents of 'Dress Code Standards.pdf'

4.2.7 Implementing various other source data processing

In the previous sections, the implementation codes were described primarily for text document files such as PDFs. Now, let's move on to processing web-based information and YouTube videos as sources of data. This will be accomplished using the open-source LLM and another vector database called FAISS, providing diverse cases for demonstration.

1) Web-Based Data Processing

When dealing with source data from web pages like company websites, product descriptions, or corporate portals, the open-source LLM GPT4All is an excellent choice. Based on the LLaMA 7B model, GPT4All offers the advantage of running on CPUs without the need for GPUs. Fig 18 demonstrates the use of GPT4All for embedding and employing LLM to process source data from web pages and store it in the vector database.

```
1 !pip install gpt4all   #GPT4All  LLM
2 !pip install langchain
3 !pip install chromadb
4
5 from langchain.document_loaders import WebBaseLoader   #WebBase Data Loader
6 loader = WebBaseLoader("https://www.samsungsds.com/us/blog/intelligent_rpa.html")
7 data = loader.load()
8
9 from langchain.text_splitter import RecursiveCharacterTextSplitter
10 text_splitter = RecursiveCharacterTextSplitter(chunk_size=500, chunk_overlap=0)
11 all_splits = text_splitter.split_documents(data)
```

```
1 from langchain.vectorstores import Chroma
2 from langchain.embeddings import GPT4AllEmbeddings   #GPT4All Embedding
3
4 vectorstore = Chroma.from_documents(documents=all_splits, embedding=GPT4AllEmbeddings())
```

```
100%|████████████████| 45.5M/45.5M [00:00<00:00, 213MiB/s]
Model downloaded at:  /root/.cache/gpt4all/ggml-all-MiniLM-L6-v2-f16.bin
```

```
1 question = "What tasks are domestic financial companies actively introducing RPA for?"
2 docs = vectorstore.similarity_search(question)
3 len(docs)
4 docs[0]
```

```
Document(page_content='In fact, banks and card companies can see instant improvements in their
work as the RPA handles so many processes, document and regulation reviews, and complex
calculations. The lists of RPA solutions in the financial sector include automatic handling of
loan documents issuance (proxy), utility bill cash disbursement, market price registration for
calculating the maximum used car loan, documentation work for settlements of international
transactions, and checking any regulation breach and', metadata={'description': 'RPA will evolve
into the complete form of AI in the future, and Cognitive Automation can play a practical role
in the evolution of RPA in the process. In other words, the existing RPA does the automated
tasks across multiple and complex systems based on predefined rules.', 'language': 'en',
'source': 'https://www.samsungsds.com/us/blog/intelligent_rpa.html', 'title': 'The Evolution of
RPA with AI | USA'})
```

**Fig 18** Web Page Data Processing (LLM - GPT4All)

2) Youtube Data Processing

In real business scenarios, it's common for companies to share promotional materials, educational content, and other resources on platforms like YouTube. To facilitate the utilization of such content, the open-source vector database FAISS can be employed. This enables the retrieval of information from YouTube videos and answering questions based on the content found within. Fig 19 showcases the implementation using FAISS to achieve this functionality.

```
1  !pip install -q openai tiktoken langchain
2  !pip install -q youtube-transcript-api youtube_search #Youtube data
3  !pip install -q faiss-cpu #Vector DB
4  from langchain.llms import OpenAI
5  from langchain.document_loaders import YoutubeLoader #Youtube Loader
6  from langchain.embeddings import OpenAIEmbeddings
7  from langchain.vectorstores import FAISS #Vector store
8  from langchain.chains import RetrievalQA
9  import tiktoken   #tokeniser for use with OpenAI's models.
10
11 embeddings = OpenAIEmbeddings()
12 loader = YoutubeLoader.from_youtube_url("https://www.youtube.com/watch?v=jDpyUusoaBg")
13 documents = loader.load()
14 db = FAISS.from_documents(documents, embeddings)
15 retriever = db.as_retriever()
16
17 llm = OpenAI(model_name="gpt-3.5-turbo", temperature=0)
18 qa = RetrievalQA.from_chain_type(llm,retriever=db.as_retriever())
19
20 query = "what age did Son Heung-min leave his country, what time does he sleep, and his thoughts on success?"
21 qa({"query": query})

{'query': 'what age did Son Heung-min leave his country, what time does he sleep, and his thoughts on success?',
 'result': "Son Heung-min left his country when he was 15 years old. He sleeps between 9 and 10 hours every day. He believes that success doesn't always come easily and that hard times make a person stronger."}
```

**Fig 19** Youtube Data Processing (VectorDB - FAISS)

4.2.8  Summary of implementation cases and expected effects

The provided implementations using Python, LangChain, OpenAI API, and the open-source ChromaDB offer a meaningful demonstration of how to easily integrate generative AI into business operations using LLM. Additionally, transitioning from a development environment to a production environment raises several key considerations and areas for improvement.

One crucial consideration is the choice of a vector database. While ChromaDB was used in the presented cases, there are numerous vector databases available, each with varying features, performance, and scalability. It's essential to carefully evaluate and select a database that suits the specific use case for optimal performance and efficiency. Furthermore, rigorous testing is vital to ensure that LLM responses remain within expected bounds and align with business requirements. This might involve improving the quality of input source data for the generative AI service, fine-tuning model parameters or system prompts to achieve better response accuracy.

In conclusion, the generative AI service built in this implementation showcases the potential of AI-powered support chatbots using internal documents. By referencing this implementation and the considerations mentioned earlier, the process of selecting technical components and integrating generative AI into business operations is expected to become smoother. It is also anticipated to contribute to the development of supporting solutions in this domain.

## V. Conclusion and Discussion

In this study, we presented methods and implementation cases for developing generative AI services using LLM application architecture, aiming to explore avenues for advancing the development and industrial utilization of generative AI technology.

We examined the theoretical background of LLM and generative AI, enhancing the understanding of both the concept of generative AI technology and the characteristics of LLM models. We also discussed ways to overcome the information scarcity challenge of LLM, either through fine-tuning or direct document information utilization, and delved into the specific functioning and key stages of the RAG model. Furthermore, we explored methods for storing and retrieving information using vector databases. By utilizing the latest LLM models and vector database technologies, we showcased various practical implementation cases applying the RAG model in business contexts. Through these cases, we confirmed that the RAG model operates by retrieving information through search capabilities and supplementing generated results, thereby providing more accurate and valuable information. This study contributes to increasing the understanding of stakeholders who intend to utilize generative AI technology within their domains, especially where research material for implementing LLM-based service components is limited.

Although this study demonstrated the implementation of generative AI services using LLM for business internal information, several limitations persist. Firstly, due to the size and complexity of LLM models, model training and implementation could consume significant time and resources. Secondly, the consistency and appropriateness of RAG model-generated results may vary, and the challenge of information scarcity might still arise. Thirdly, while open-source-based implementations were provided in most cases, certain functional aspects might be lacking.

Future research should address these limitations and seek more efficient utilization of LLM and RAG models. Firstly, research efforts should focus on reducing the size and complexity of LLM models, exploring methods to achieve effective results with smaller and simpler models, such as sLLM. Secondly, new approaches that enhance the search capabilities of the RAG model and address information scarcity challenges could be developed to apply more accurate and efficient information retrieval techniques, thereby enhancing model performance. Thirdly, finding ways to improve the consistency and appropriateness of RAG model-generated results is essential. Exploring and applying various techniques to enhance the quality of generated content is important. Additionally, consideration should be given to using high-performance LLM service component solutions that are commercially available for different business scenarios to enhance industrial viability.

Lastly, it is vital to implement LLM and RAG models for generative AI services in various languages and cultural contexts. This endeavor will promote the advancement and industrial utilization of generative AI technology in diverse regions. The combined efforts of these initiatives are expected to expand the scope of generative AI technology utilization, enabling practical applications in the real world.